# Synergistic Reconstruction and Synthesis via Generative Adversarial Networks for Accelerated Multi-Contrast MRI

Salman Ul Hassan Dar[1,2], Mahmut Yurt[1,2], Mohammad Shahdloo[1,2],

Muhammed Emrullah Ildız[1,2], Tolga Çukur[1,2,3]

[1]Department of Electrical and Electronics Engineering, Bilkent University, Ankara, Turkey
[2]National Magnetic Resonance Research Center (UMRAM), Bilkent University, Ankara, Turkey
[3]Neuroscience Program, Sabuncu Brain Research Center, Bilkent University, Ankara, Turkey

*Running title:* Synergistic Reconstruction and Synthesis for Accelerated Multi-Contrast MRI

*Address correspondence to:*

Tolga Çukur

Department of Electrical and Electronics Engineering, Room 304
Bilkent University
Ankara, TR-06800, Turkey

TEL: +90 (312) 290-1164

E-MAIL: cukur@ee.bilkent.edu.tr

This work was supported in part by a European Molecular Biology Organization Installation Grant (IG 3028), by a TUBA GEBIP fellowship, and by a BAGEP fellowship awarded to T. Çukur. We gratefully acknowledge the support of NVIDIA Corporation with the donation of the Titan GPUs used in this study.

**Abstract**

Multi-contrast MRI acquisitions of an anatomy enrich the magnitude of information available for diagnosis. Yet, excessive scan times associated with additional contrasts may be a limiting factor. Two mainstream approaches for enhanced scan efficiency are reconstruction of undersampled acquisitions and synthesis of missing acquisitions. In reconstruction, performance decreases towards higher acceleration factors with diminished sampling density particularly at high-spatial-frequencies. In synthesis, the absence of data samples from the target contrast can lead to artefactual sensitivity or insensitivity to image features. Here we propose a new approach for synergistic reconstruction-synthesis of multi-contrast MRI based on conditional generative adversarial networks. The proposed method preserves high-frequency details of the target contrast by relying on the shared high-frequency information available from the source contrast, and prevents feature leakage or loss by relying on the undersampled acquisitions of the target contrast. Demonstrations on brain MRI datasets from healthy subjects and patients indicate the superior performance of the proposed method compared to previous state-of-the-art. The proposed method can help improve the quality and scan efficiency of multi-contrast MRI exams.

**Keywords:** generative adversarial network, synthesis, reconstruction, multi-contrast MRI, variable density.

# 1 – Introduction

Magnetic resonance imaging (MRI) is a preferred modality for assessment of soft tissues due the diversity of contrasts that it can provide. A typical MRI protocol comprises a set of pulse sequences that capture images of the same anatomy under different contrasts, with the aim to enhance diagnostic information. For instance, in neuroimaging protocols, $T_1$-weighted images are useful for delineation of gray and white matter, whereas $T_2$-weighted images are more useful for delineation of fluids and fat. Although acquisition of multiple distinct contrasts is desirable, it may not be feasible due to scan time limitations or uncooperative patients. Thus, methods for accelerating MRI acquisitions without compromising image quality are of great interest for multiple-contrast applications.

The predominant approach for accelerated MRI relies on undersampled k-space acquisitions for scan time reduction, and on reconstruction algorithms for recovery of missing samples based on the collected evidence (i.e., acquired samples) [1]–[5]. Given the compressible nature of MR images, the state-of-the-art framework for achieving this recovery is compressive sensing (CS) [3], [4]. CS methods commonly employ variable-density random undersampling in k-space to capture most of the energy in the MR image while ensuring low coherence of aliasing artifacts. The inverse problem of image reconstruction from sub-Nyquist sampled data is then solved with the help of additional regularization terms. These terms reflect prior information based on the assumption that MR images are sparse or compressible in a known transform domain (e.g., wavelet, total variation). Recent CS methods have further improved recovery performance by incorporating dictionary learning methods or deep neural networks to adaptively learn the sparsifying transform domains [6]–[14]. Despite the promise of CS-MRI, however, the evidence collected on the target MR image diminishes towards high acceleration factors. In turn, this degrades the recovery performance, and causes loss in particularly high-spatial-resolution image features that may be relevant for diagnosis.

An entirely different approach to accelerated MRI is to perform fully-sampled acquisitions of a subset of the desired contrasts, and then to synthesize images of the missing contrast based on the collected one. Synthesis requires a learning-based framework where an intensity-based mapping between the target contrast and the source contrast is estimated using a collection of training image pairs in both contrasts, and then applied on test images [15]–[32]. A common learning method is based on compressive sensing that relies on the assumption that target patches can be expressed as sparse linear combinations of patches from the training images [24]. Recent synthesis methods have proposed architectures based on neural networks to learn direct nonlinear mappings between the source and target contrasts with enhanced accuracy [17], [26]–[28], [31], [32]. Note that, in contrast to CS-MRI, synthesis methods have access to high-spatial-frequency information in the fully-sampled source images, and to the extent that this information is shared across contrasts it can yield improved immunity against loss of spatial resolution in the target contrast. Yet, local inaccuracies may occur in synthesized images when the source contrast is less sensitive to differences in relaxation parameters of two tissues compared to the target contrast, or vice versa. For instance, inflammation can be more clearly delineated from normal tissues in $T_2$-weighted as opposed to $T_1$-weighted images. In such cases, synthesized images might contain artificial pathology or fail to depict existing pathology.

Here we propose a new approach that that synergistically merges the compressed-sensing and synthesis for enhanced performance in accelerated MRI. The proposed approach takes as input either fully-sampled or lightly undersampled acquisitions of the source contrast, and heavily undersampled acquisitions from one or more target contrasts. To recover images of the target contrast, it simultaneously leverages the relatively low-spatial-frequency information available in the collected evidence for the target contrast and the relatively high-spatial frequency information available in the source contrast. The input-to-output mapping is implemented using generative adversarial networks (GAN) that were recently shown to outperform traditional methods for image reconstruction and synthesis tasks [6]–[8], [13], [32]. The proposed reconstructing-synthesizing GAN (rsGAN) contains a generator network for estimating the target-contrast image given linear reconstructions of the undersampled images in the source and target contrasts; and a discriminator network to ensure that recovered images are as realistic as possible [33].

We demonstrated the proposed approach on two separate datasets containing normal subjects, and one dataset containing patients suffering from high- or low-grade glioma. Two competing methods were considered: a pure reconstructing network (rGAN) that recovers the target-contrast image given undersampled images of the target contrast, and a pure synthesizing network (sGAN) that synthesizes the target-contrast image given undersampled images of the source contrast. Our results indicate that, rsGAN yields enhanced performance compared to both rGAN and sGAN. In particular, rsGAN enables higher acceleration factors compared to rGAN since it more reliably recovers high-spatial-frequency information. Compared to sGAN, rsGAN achieves improved reliability against artificial feature loss or leakage since it uses collected evidence from the target contrast to prevent hallucination. Overall, the proposed approach can successfully recover MR images of at acceleration factors up to 50x in the target contrasts, enabling a significant improvement in multi-contrast MRI.

Following are our main contributions:

1 – To our knowledge, this is the first GAN-based architecture that jointly reconstructs and synthesizes target contrasts to accelerate multi-contrast MRI acquisitions.

2 – The proposed approach can enable high acceleration factors up to 50x by incorporating information from both source and target contrasts.

3 – The proposed approach can successfully recover pathologies that are either missing in the source contrast or are not clearly visible in the undersampled acquisitions of the target contrast.

4 – The proposed approach can jointly reconstruct and synthesize the target contrast even when the source contrasts are lightly undersampled.

## 2 – Theory and Methods

### 2.1 - Accelerated MRI

Two mainstream approaches that can be used to accelerate MR acquisitions and enhance the diversity of acquired contrasts are reconstruction of a target contrast given randomly undersampled acquisitions of the same contrast, and synthesis of a target contrast based on fully-sampled acquisitions of a distinct source contrast. Both approaches incorporate prior information about image structure to improve the conditioning of the inverse problem of recovering images of the target contrast. However, they differ fundamentally in the type of prior information used. The problem formulations for reconstruction and synthesis are overviewed below.

**Reconstruction:** In this case, MR acquisitions are accelerated commonly via variable-density random undersampling patterns:

$$F_u m_1 = y_{1a} \qquad (1)$$

where $F_u$ is the partial Fourier operator defined at the k-space sampling locations, and $m_1$ is the image of the target contrast, $y_{1a}$ are the acquired k-space data. The reconstruction task is then to recover the target image given the collected evidence (i.e., acquired data). Note that the problem in Eq. 1 is ill-posed, thus successful recovery requires additional prior information about the image. In the CS framework, this prior information reflects the sparsity of the image in a known transform domain (i.e., wavelet, TV transforms). The prior can be incorporated into the inverse problem as a regularization term:

$$\widehat{m_1} = \arg\min_{m_1} \lambda \|F_u m_1 - y_{1a}\|_2 + R(m_1) \qquad (2)$$

where the first term enforces consistency of the reconstructed and acquired data in k-space, $R(m_1)$ is the regularization term reflecting the prior, and $\lambda$ controls the relative weighting of data consistency against the prior. $R(m_1)$ typically involves the $l_0$ or $l_1$-norm of transform coefficients.

Recent studies have proposed neural-network methods to adaptively learn both nonlinear transform domains directly from MRI data and how to recover images from these domains. In the training stage, a large dataset of pairs of undersampled and fully-sampled acquisitions are leveraged to learn the network-based solution to the inverse problem:

$$\mathcal{L}_{rec}(\theta) = E_{m_{1t}^u, m_{1t}} \|G(m_{1t}^u; \theta) - m_{1t}\|_p \qquad (3)$$

where $m_{1t}^u$ and $m_{1t}$ represent undersampled and fully-sampled training images, $G(m_{1t}^u; \theta)$ is the reconstructed output of the neural network based on network parameters $\theta$, and $\|.\|_p$ denotes $l_p$-norm (where p is typically 1 or 2). Once the network parameters that minimize the objective in Eq. 3 have been learned, the following optimization problem can be cast to obtain reconstructions of undersampled acquisitions:

$$\widehat{m_1} = \arg\min_{m_1} \lambda \|F_u m_1 - y_{1a}\|_2 + \|G(m_1^u; \theta^*) - m_1\|_2 \qquad (4)$$

where $m_1^u$ is the undersampled image, $G(m_1^u; \theta^*)$ is the reconstruction by the trained network with parameters $\theta^*$, and $\widehat{m_1}$ is final recovered image. In Eq. 4, the first term again enforces consistency of reconstructed and acquired data. The second term is analogous to $R(m_1)$ in Eq. 2, and it enforces consistency of the recovered image to the network reconstruction.

**Synthesis:** In the synthesis case, fully-sampled images of a source contrast are assumed to be available. The task is then to recover target-contrast images ($m_1$) given source-contrast images ($m_2$) of the same anatomy. A learning-based procedure is used to estimate a mapping between the source and target contrast images. In the training stage, a large dataset of pairs of fully-sampled images from the source and target contrasts are used ($m_{2t}, m_{1t}$). In the CS-based synthesis framework, patch-based dictionaries ($\Phi_2, \Phi_1$) are formed for both source and target contrasts using $m_{2t}$ and $m_{1t}$. These dictionaries are analogous to the sparsifying transform domains used in CS reconstructions. The aim is to express each patch in the source contrast images $m_2$ as a sparse linear combination of transform coefficients of the corresponding dictionary atoms:

$$\alpha(j) = \arg\min_{\alpha(j)} \|m_2(j) - \Phi_2.\alpha(j)\|_2 + \|\alpha(j)\|_1 \quad (5)$$

where $\alpha(j)$ is the learned combination coefficients for the j$^{th}$ patch, $m_2(j)$ denotes the j$^{th}$ patch in the source contrast, and $\Phi_2$ denotes the dictionary formed using patches from $m_{2t}$. The first term ensures consistency of the synthesized patch to the true patch. The second term enforces sparsity of the vector of combination coefficients. Once the combination is learned, it can be used to synthesize target contrast images :

$$\widehat{m_1}(j) = \Phi_1.\alpha(j) \quad (6)$$

where $\Phi_1$ denotes the dictionary formed using patches from $m_{1t}$, and $\widehat{m_1}(j)$ is the j$^{th}$ patch of the final synthesized image.

Recent studies have proposed neural-network methods to directly learn an adaptive, non-linear mapping from the source contrast to the target contrast. In the training stage, network parameters are optimized based on a loss function that reflects the error between the network output and the true target image:

$$\mathcal{L}_{synth}(\theta) = E_{m_{1t},m_{2t}} \|G(m_{2t};\theta) - m_{1t}\|_p \quad (7)$$

where $m_{1t}$ and $m_{2t}$ represent pairs of source and target images, and $G(m_{2t};\theta)$ is the mapping from source to target contrast characterized by parameters $\theta$. Once the network parameters that minimize the objective in Eq. 7 have been learned, the network output can be directly calculated to obtain the synthesis results:

$$\widehat{m_1} = G(m_2;\theta^*) \quad (8)$$

where $\widehat{m_1}$ is the prediction using the mapping $G(m_2;\theta^*)$ with parameters $\theta^*$. Unlike the reconstruction task, here there is no evidence that has been collected about the target contrast. Therefore, no optimization procedures are needed for synthesis in the testing stage.

**2.2 – Joint reconstruction-synthesis via conditional GANs**

In the reconstruction task, the inverse problem solution uses undersampled acquisitions of the target contrast as evidence, and intrinsic image properties such as sparsity as prior. As the acceleration factor grows, evidence becomes scarce particularly towards high spatial frequencies that are sparsely covered by variable-density patterns. This in turn elevates the degree of aliasing artifacts; and if heavier weighting is given to the prior as a remedy, important features may be lost in the recovered images. Meanwhile, in the synthesis task, the inverse problem solution uses fully-sampled acquisitions of a distinct source contrast of the same anatomy as a prior. When the source and target contrasts exhibit similar levels of sensitivity to differences in tissue parameters, this prior can enable successful solution of the inverse problem. However, when the source and target show differential sensitivity, then features that are not supposed to be in the target may leak from the source onto the synthesized image, or features that must be present in the target may be missed.

To address the limitations of pure reconstruction or synthesis, we proposed to synergistically combine the two approaches with the aim to enhance recovery of multi-contrast MRI images. The proposed approach allows for fully-sampled or lightly undersampled acquisitions of the source contrasts, and lightly to heavily undersampled acquisitions of the target contrasts. Given $k$ target contrasts and $n-k$ source contrasts, the joint recovery problem can be formulated as:

$$\widehat{m}_{1,2,3,\ldots,n} = \min_{m_{1,2,\ldots,n}} \lambda \sum_{i=1}^{k} \left\| F_u m_i^{hu} - y_{ia} \right\|_2 + \lambda \sum_{j=k+1}^{n} \left\| F_u m_j^{lu} - y_{ja} \right\|_2 + R\left(m_1^{hu}, \ldots, m_k^{hu}, m_{k+1}^{lu}, \ldots, m_n^{lu}\right) \quad (9)$$

where $R\left(m_1^{hu}, \ldots, m_k^{hu}, m_{k+1}^{lu}, \ldots, m_n^{lu}\right)$ is a regularization term based on prior information, $m_i^{hu}$ is the i[th] contrast that is heavily undersampled (i.e., target contrast), and $m_j^{lu}$ is j[th] contrast that is lightly undersampled (i.e., source contrast), and $y_{ia}$ denotes the acquired data for the i[th] contrast. We recast Eq. 9 using a neural-network based formulation:

$$\widehat{m}_{1,2,3,\ldots,n} = \min_{m_{1,2,\ldots,n}} \lambda \sum_{i=1}^{k} \left\| F_u m_i^{hu} - y_{ia} \right\|_2 + \lambda \sum_{j=k+1}^{n} \left\| F_u m_i^{lu} - y_{ja} \right\|_2 + \sum_{l=1}^{n} \left\| G\left(m_1^{hu}, \ldots, m_k^{hu}, m_{k+1}^{lu}, \ldots, m_n^{lu}; \theta^*\right)[l] - m_l \right\|_2 \quad (10)$$

Here, multiple separate channels for network output are considered since multiple contrast images can be recovered simultaneously. In Eq. 10, $G\left(m_1^{hu}, \ldots, m_k^{hu}, m_{k+1}^{lu}, \ldots, m_n^{lu}; \theta^*\right)[l]$ denotes the l[th] channel of the network output, among a total of n channels for the entire set of contrasts. The first two terms respectively enforce the consistency of reconstructed data to acquired data in the target and source contrasts. The last term enforces consistency of the network outputs to the recovered images. Solution of Eq. 10 yields estimates of the images for each contrast separately as:

$$y_{ir}(k) = \begin{cases} \dfrac{F\{G\left(m_1^{hu}, \ldots, m_k^{hu}, m_{k+1}^{lu}, \ldots, m_n^{lu}; \theta^*\right)[i]\}(k) + \lambda y_{ia}(k)}{1+\lambda}, & \text{if } k \in \Omega \\ F\{G\left(m_1^{hu}, \ldots, m_k^{hu}, m_{k+1}^{lu}, \ldots, m_n^{lu}; \theta^*\right)[i]\}(k), & \text{otherwise} \end{cases} \quad (11)$$

$$\widehat{m}_i = F^{-1}\{y_{ir}\}$$

where $y_{ir}$ denotes the k-space representation of the image for the i[th] contrast, $\Omega$ is the set of acquired k-space samples, $F$ is the Fourier transform operator, and $F^{-1}$ is the inverse Fourier transform operator. The solution stated above performs two subsequent projections on the input images. The first projection takes undersampled acquisitions to generate the network predictions. The second projection enforces data consistency between data sampled that were originally acquired and those that are predicted by the network.

Based on the recent progress by generative adversarial networks in MR image synthesis and reconstruction tasks, we chose to build the joint recovery network using a conditional GAN architecture. Our network contains two subnetworks: a generator and a discriminator. The task of the generator is to learn a mapping from undersampled acquisitions onto fully-sampled acquisitions of source and target images. Meanwhile, the task of the discriminator is to differentiate between the images predicted by the generator and the actual images. As such, an adversarial loss function is typically used to train both subnetworks:

$$\mathcal{L}_{Adv}(\theta_D, \theta_G) = E_{\boldsymbol{m_t}}[\log D(\boldsymbol{m_t}; \theta_D)] + E_{\boldsymbol{m_t},z}[\log(1 - D(\boldsymbol{m_t}, G(z; \theta_G)))] \quad (12)$$

where $\boldsymbol{m_t}$ represent the MR images aggregated across $n$ contrasts $(m_1, m_2, ..., m_n)$ in the training dataset, $G$ is the generator with parameters $\theta_G$, $D$ is the discriminator with parameters $\theta_D$, $z$ is the latent variable and $\mathcal{L}_{Adv}(\theta_D, \theta_G)$ is the adversarial loss function. The discriminator tries to maximize while the generator tries to minimize the adversarial loss. Ideally the generator should be able to generate images that the discriminator cannot separate from real images. Here, to prevent vanishing gradients problems, we used a modified loss as in LSGAN [34]:

$$\mathcal{L}_{Adv}(\theta_D, \theta_G) = E_{\boldsymbol{m_t}}[(D(\boldsymbol{m_t}; \theta_D) - 1)^2] + E_z[D(G(z; \theta_G))^2] \tag{13}$$

Note that the images of the target contrast are statistically dependent on both collected evidence in the target contrast and well as images of the same anatomy under the source contrast. To better capture these dependencies, a conditional GAN model was considered that takes as input both undersampled acquisitions of the source and target contrasts [35]. The loss function in Eq. 13 is then reformulated as:

$$\mathcal{L}_{condAdv}(\theta_D, \theta_G) = E_{\boldsymbol{m_t}}[(D(\boldsymbol{m_t}; \theta_D) - 1)^2] + E_{\boldsymbol{m_t^{hu}}, \boldsymbol{m_t^{lu}}}[D(G(\boldsymbol{m_t^{hu}}, \boldsymbol{m_t^{lu}}; \theta_G))^2]] \tag{14}$$

where $\boldsymbol{m_t^{hu}}$ represents the heavily undersampled acquisitions aggregated across $k$ target contrasts $(m_1, m_2, ..., m_k)$, and $\boldsymbol{m_t^{lu}}$ represents the lightly undersampled acquisitions aggregated across $n - k$ source contrasts $(m_{k+1}, m_{k+2}, ..., m_n)$. The latent variable $z$ was omitted here because it was observed to have insignificant effect on the network performance. Since multiple images are to be recovered by the network model, an independent output channel was assigned to each contrast. To ensure reliable recovery in each channel, a pixel-wise loss function was incorporated to the generator:

$$\mathcal{L}_{L1}(\theta_G) = E_{\boldsymbol{m_t}, \boldsymbol{m_t^{hu}}, \boldsymbol{m_t^{lu}}}[\|G(\boldsymbol{m_t^{hu}}, \boldsymbol{m_t^{lu}}; \theta_G) - \boldsymbol{m_t}\|_1] \tag{15}$$

The adversarial and pixel-wise losses were finally combined to train the proposed reconstructing-synthesizing GAN (rsGAN) model:

$$\mathcal{L}_{rsGAN}(\theta_D, \theta_G) = \lambda_p \mathcal{L}_{L1}(\theta_G) + \mathcal{L}_{condAdv}(\theta_D, \theta_G) \tag{16}$$

where $\lambda_p$ is the relative weighting of the pixel-wise loss function.

### 2.3 – Competing methods

To evaluate the effectiveness of rsGAN, we compared it against to other GAN architectures. The first network was trained to only perform synthesis of the target-contrast images based on the respective source-contrast images. Source-contrast images were taken to be fully-sampled, high-quality images. We will refer to this network as the synthesizing GAN (sGAN). The second network was trained to only perform reconstruction of the target-contrast images based on undersampled acquisitions of the target contrast. For each target contrast a separate network was trained to perform reconstruction. We will refer to this network as the reconstructing GAN (rGAN).

### 2.4 – MRI Datasets

We demonstrated the proposed approach on three different public datasets containing multi-contrast MRI images. The public datasets MIDAS [36] and IXI (*http://brain-development.org/ixi-dataset/*) comprised images collected in healthy normals. BRATS (*https://sites.google.com/site/braintumorsegmentation/home/brats2015*) comprised images collected in patients with low-grade glioma (LGG) or high-grade glioma (HGG). Relevant details about each dataset are given below.

**MIDAS dataset:** $T_1$-weighted and $T_2$-weighted images in the MIDAS dataset were considered. Data from 62 subjects were analyzed, where 47 subjects were used in the training stage and 15 subjects were reserved for the testing stage. Within each volumetric image set in individual subjects, approximately 100 central axial cross-sections that contained brain tissues and that were relatively free of aliasing artifacts were manually selected. The scan protocols were as follows: for $T_1$-weighted images, 3D gradient-echo sequence, repetition time (TR)=14ms, echo time (TE)=7.7ms, flip angle=$25^0$, volume size=256x176x256, voxel dimensions=1mm×1mm×1mm; for $T_2$-weighted images, 2D spin-echo sequence, repetition time (TR)=7730ms, echo time (TE)=80ms, flip angle=$180^0$, volume size=256x192x256, voxel dimensions=1mm×1mm×1mm.

**IXI dataset:** $T_1$-weighted, $T_2$-weighted and PD-weighted images in the IXI dataset were considered. Data from 37 subjects were analyzed, where 28 subjects were used in the training stage and 9 subjects were reserved for the testing stage. Within each volumetric image set in individual subjects, approximately100 axial cross-sections that contained brain tissues and that were free of artifacts were manually selected. The scan protocols were as follows: for $T_1$-weighted images, repetition time (TR)=9.813ms, echo time(TE)=4.603ms, flip angle=$8^0$, volume size=256×256×150, voxel dimensions=0.94mm×0.94mm×1.2mm; for $T_2$-weighted images, repetition time (TR)=8178ms, echo time(TE)=100ms, flip angle=$90^0$, volume size=256×256×130, voxel dimensions= 0.94mm×0.94mm×1.2mm; for PD-weighted images, Repetition time (TR)=8178ms, echo time(TE)=8ms, flip angle=$90^0$, volume size=256×256×130, voxel dimensions=0.94mm×0.94mm×1.2mm.

**BRATS dataset:** $T_1$-weighted and $T_2$-weighted images in the BRATS dataset were considered. Data from 40 subjects were analyzed, where 30 subjects were used in the training stage and 10 subjects were reserved for the testing stage. Within each volumetric image set in individual subjects, approximately 100 central axial cross-sections that contained brain tissues and that were relatively free of aliasing artifacts were manually selected. Since the data were acquired in various different sites, no single scan protocol existed.

### 2.5 - Image registration

Since the images in the MIDAS and IXI datasets were unregistered, these images were registered before training and testing. For the MIDAS dataset, $T_2$-weighted images of each subject were registered onto $T_1$-weighted images of the same subject using a rigid transformation. Images were registered based on mutual information loss. For the IXI dataset $T_2$- and PD-weighted images of each subject were registered onto $T_1$-weighted images of each subject using an affine transformation. Images were registered based on mutual information loss. Registrations were carried out using FSL [37], [38].

### 2.6 - Undersampling Patterns

For heavily undersampled acquisitions of the target contrast, we examined acceleration factors in a broad range R=10x, 20x, 30x, 40x, 50x. For lightly undersampled acquisitions of the source contrast, we examined acceleration factors in a relatively limited range R=1x, 2x, 5x. Sampling patterns were generated using the variable-density Poisson disc sampling method [5]. Fully-sampled images were Fourier transformed, and then retrospectively sampled using the generated patterns. For each dataset, a total of 100 distinct random patterns were generated for use during the training stage, and a separate set of 100 random patterns were generated for use during the testing stage. The sampling density profiles across k-space varied for different values of R. The density profiles were designed based on a polynomial function with the following radius of fully-sampled calibration region ($k_r$) and polynomial degree (d): for 2x≤R≤10x, $k_r$ = 0.14, d = 5; for R=20x, $k_r$ = 0.10, d = 9; for R=30x, $k_r$ = 0.10, d = 10; for R=40x, $k_r$ = 0.10, d = 15; for R=50x, $k_r$ = 0.06, d = 20.

### 2.7 – Model Training Procedures

All GAN-based models (rsGAN, sGAN and rGAN) were trained using an identical set of procedures.

To train each conditional GAN, we adopted the generator and discriminator from [39], [40]. The generator consisted of the following layers connected in series: Convolution layer (kernel-size=7, output-features=64, stride=1, activation=ReLU), convolutional layer (kernel-size=3, output-features=64, stride=2, activation=ReLU), convolutional layer (kernel-size=3, output-features=256, stride=2, activation=ReLU), 9x resnet blocks (kernel-size=3, output-features=256, stride=1, activation=ReLU), fractionally-strided convolutional layer (kernel-size=3, output-features=128, stride=2, activation=ReLU), fractionally-strided convolutional layer (kernel-size=3, output-features=64, stride=2, activation=ReLU), up-sampling convolutional layer (kernel-size=3, output-features=128, stride=2, activation=ReLU), convolutional layer (kernel-size=7, output-features=1, stride=1, activation=tanh). The discriminator consisted of the following layers connected in series: Convolution layer (kernel-size=4, output-features=64, stride=2, activation=leakyReLU), Convolution layer (kernel-size=4, output-features=128, stride=2, activation=leakyReLU), Convolution layer (kernel-size=4, output-features=256, stride=2, activation=leakyReLU), Convolution layer (kernel-size=4, output-features=512, stride=1, activation=leakyReLU), Convolution layer (kernel-size=4, output-features=1, stride=1, activation=none).

Generator and discriminator networks were trained for 200 epochs using the Adam optimizer [41], with decay rates for the first and second moment estimates set as 0.5 and 0.999. For the generator, the learning rate was set as 0.0002 for the initial 100 epochs and then linearly decayed to 0. For the discriminator, the learning rate was set as 0.0001 for the first 100 epochs and then linearly decayed to 0 during the remaining epochs. Dropout regularization was used to enhance the generalizability of the network model, with a dropout rate of 0.5. Instance normalization was applied [42]. All model weights were randomly initialized based on a normally-distributed variable with 0 mean and 0.02 standard deviation. Relative weighting of the pixel-wise loss function against the adversarial loss function ($\lambda_p$) was set to 100. Relative weighting of data consistency against the prior ($\lambda$) was set to infinity. Note that the networks receive as inputs Fourier reconstructions of undersampled acquisitions that are complex valued. For each input contrast, two channels were designated to represent the real and imaginary image components.

During the testing phase, for the case where the source contrasts were lightly undersampled, first a network was trained to reconstruct source-contrast images from undersampled data of the source contrast. The reconstructed images were then fed to rsGAN that was trained to synergistically reconstruct-synthesize images of the target contrasts from fully sampled images of the source contrasts and highly undersampled images of the target contrasts, and sGAN that was trained to synthesize images of the target contrasts from fully sampled images of the source contrasts.

**2.8 – Experiments**

To evaluate the comparative performance of the proposed approach, rsGAN, rGAN and sGAN were individually trained and tested on multi-contrast MRI datasets. Theoretically, as R approaches 1x, rsGAN and rGAN should show nearly identical performance that is superior to sGAN since sGAN has no evidence collected about the target contrast. As R goes to infinity, rsGAN and sGAN should show nearly identical performance that is superior to rGAN, since no evidence from the target contrast will be available to any of the networks. In intermediate R values, we reasoned that rsGAN would outperform rGAN in terms of reliability in recovery of high-frequency information since variable-density patterns suboptimally sample high spatial frequencies in the target contrast. We also reasoned that rsGAN would outperform sGAN especially when the source and target contrasts showed differential sensitivity to differences in tissue parameters. Based on these notions, we measured the performance of all three methods across a broad range of acceleration factors.

In both MIDAS and BRATS datasets, we considered two main scenarios. First, $T_1$-weighted acquisitions were taken as the source contrast (R=1x), and $T_2$-weighted acquisitions were taken as the target contrast (R=10x, 20x, 30x, 40x, 50x). Second, $T_2$-weighted acquisitions were taken as the source (R=1x), and $T_1$-weighted acquisitions were taken as the target (R=10x, 20x, 30x, 40x, 50x).

Two distinct scenarios were examined in the IXI dataset. First, $T_1$-weighted acquisitions were taken as the source contrast (R=1x), and both $T_2$- and PD-weighted acquisitions were taken as the target contrasts (R=10x, 20x, 30x, 40x, 50x). Since $T_2$-and PD-weighted acquisitions are typically performed using similar sequences, the acceleration factors for these two contrasts were always matched. Second, the source $T_1$-weighted acquisitions were lightly undersampled (R=2x, 5x), and $T_1$-, $T_2$-, and PD-weighted images were jointly recovered.

All network models and conventional reconstruction and synthesis techniques were trained and tested on the same instances of data and undersampling patterns. To quantitatively assess the quality of recovered images, the fully-sampled reference images were used. All images were first normalized to the range [0 1]. Then peak signal-to-noise ratio (PSNR) and structural similarity index measure (SSIM) were calculated between the recovered and reference images. Statistical significance of differences in PSNR and SSIM between methods were assessed via a nonparametric Wilcoxon signed-rank test.

## 3 – Results

To demonstrate the proposed approach, rsGAN, rGAN and sGAN were individually trained and tested on multi-contrast MRI datasets for a broad range of acceleration factors (R). We first considered two separate models on the MIDAS dataset: a model to recover $T_2$-weighted images given $T_1$-weighted images as source contrast, and another to recover $T_1$-weighted images given $T_2$-weighted images as source contrast. Tables I and II list the respective PSNR measurements for each model, and Fig. 2 illustrates performance as a function of R. In $T_2$-weighted image recovery, rsGAN outperforms both rGAN and sGAN in all examined cases ($p<0.01$), except for rGAN at R=10x where the competing methods perform similarly. Overall, rsGAN achieves 1.73±1.01 dB (mean±std across R) higher PSNR and 4.4±2.6 % higher SSIM than rGAN, and 3.25±1.62 dB higher PSNR and 4.0±2.3 % higher SSIM than sGAN. In $T_1$-weighted image recovery, rsGAN outperforms both rGAN and sGAN in all examined cases ($p<0.01$), except for rGAN at R=20x where rGAN performs similarly in terms of SSIM. Overall, rsGAN achieves 1.10±0.63 dB higher PSNR and 4.8±3.2 % higher SSIM than rGAN, and 2.96±1.01 dB higher PSNR and 5.2±2.0 % higher SSIM than sGAN.

$T_2$- and $T_1$-weighted images in the MIDAS dataset recovered while R is varied from 10x to 50x are displayed in Figs. 3 and 4, respectively. Representative $T_2$- and $T_1$-weighted images recovered with ZF, rGAN, sGAN and rsGAN at R=50x are shown in Fig. 5. As expected, the similarity between rsGAN and rGAN results increases towards R=10x, and rsGAN and that between rsGAN and sGAN increases towards R=50x. That said, rsGAN recovers images of higher visual quality and acuity than both competing methods, particularly at intermediate R values. These results indicate that the incorporation of a fully-sampled acquisitions of the source contrast enables rsGAN to more reliably recover high-frequency information compared to rGAN, and that the use of evidence collected on the target contrast ensures that rsGAN yields more accurate recovery compared to sGAN.

Next, we demonstrated the proposed method on a dataset acquired in patients with high- or low-grade gliomas. We considered two models on the BRATS dataset: a model to recover $T_2$-weighted images given $T_1$-weighted images, and another to recover $T_1$-weighted images given $T_2$-weighted images. Tables III and IV list the respective PSNR values, and Fig. 6 illustrates model performance as a function of R. In $T_2$-weighted image recovery, rsGAN outperforms both rGAN and sGAN in all examined cases ($p<0.01$), except for rGAN at R=10x and 20x where the competing methods perform similarly. Overall, rsGAN achieves 1.12±0.87 dB higher PSNR and 3.9±3.5 % higher SSIM than rGAN, and 6.14±1.87 dB higher PSNR and 3.4±3.7 % higher SSIM than sGAN. In $T_1$-weighted image recovery, rsGAN competing methods ($p<0.01$), except for rGAN at R=20x where rGAN performs similarly in terms of PSNR. Overall, rsGAN achieves 1.78±0.84 dB higher PSNR and 4.8±3.2 % higher SSIM than rGAN, and 5.83±1.71 dB higher PSNR and 2.3±3.0 % higher SSIM than sGAN.

Representative $T_2$- and $T_1$-weighted images in the BRATS dataset recovered with ZF, rGAN, sGAN and rsGAN at R=50x are shown in Fig. 7. Similar to the assessment on the previous dataset with normal subjects, rsGAN recovers images of higher visual quality and acuity than competing methods, particularly at intermediate R values. Note that multi-contrast images can show differential sensitivity to tumor tissue, where tumors can be more easily delineated in $T_2$- versus $T_1$-weighted images particularly in patients with low-grade glioma. As a result, sGAN suffers from either loss of features in the target contrast (e.g., during recovery of $T_2$-weighted images) or synthesis of artefactual features (e.g., during recovery of $T_1$-weighted images). Meanwhile, rGAN suffers from excessive loss of high spatial frequency information at high R. In comparison, rsGAN achieves higher spatial acuity while preventing feature losses and artefactual synthesis. Thus, the rsGAN method enables more reliable and accurate recovery when the source contrast is substantially less or more sensitive to differences in relaxation parameters of two tissues compared to the target contrast.

Lastly, we demonstrated the utility of rsGAN to recover multiple target contrasts simultaneously. The specific model tested on the IXI dataset was aimed to recover both $T_2$- and PD-weighted images given

$T_1$-weighted images as source contrast. We examined the effect of light undersampling performed on the source contrast ($R_{T1}$=1x, 2x, 5x) in addition to heavy undersampling on the target contrasts (R=10x, 20x, 30x, 40x, 50x). Tables V and VI list the PSNR and SSIM measurements for $T_2$- and PD-weighted images, respectively. Figure 8 illustrates model performance as a function of $R_{T1}$ and R. In $T_2$-weighted recovery, rsGAN ($R_{T1}$=1x) outperforms rGAN in all examined cases (p<0.01). The rsGAN method also outperforms sGAN in terms of PSNR at all R (p<0.01). Overall, rsGAN achieves 2.84±1.01 dB higher PSNR and 10.1±4.7 % higher SSIM than rGAN, and 2.88±1.69 dB higher PSNR than sGAN. In PD-weighted image recovery, rsGAN again outperforms rGAN in all examined cases, and it outperforms sGAN in terms of PSNR at all R (p<0.01). Overall, rsGAN achieves 2.22±0.58 dB higher PSNR and 11.0±4.7 % higher SSIM than rGAN, and 2.23±1.63 dB higher PSNR than sGAN. In the IXI dataset, a comparison of SSIM values of rsGAN versus sGAN does not yield consistent results. As expected, the performance of rsGAN gradually decreases for higher values of $R_{T1}$. However, even at $R_{T1}$=5 rsGAN outperforms both rGAN and sGAN in terms of PSNR across all R.

Representative $T_2$- and PD-weighted images in the IXI dataset recovered with ZF, rGAN, sGAN and rsGAN at $R_{T1}$=2x, R=50x are shown in Fig. 9. The rsGAN method yields sharper images and improved suppression of aliasing artifacts compared to rGAN and sGAN. Importantly, these improvements are apparent even when the source contrast acquisitions are accelerated.

# 4 – Discussion

A synergistic reconstruction-synthesis approach based on conditional GANs was presented for highly accelerated multi-contrast MRI. In this approach, several source- and target-contrast acquisitions accelerated to various degrees are taken as input, and high-quality images for individual contrasts are then recovered. The proposed rsGAN method yielded superior recovery performance against state-of-the-art reconstruction and synthesis methods in three public MRI datasets. While rsGAN was demonstrated for multi-contrast MRI here, it may also offer improved performance in recovery of images in accelerated multi-modal datasets.

Several previous studies considered joint reconstructions of multi-contrast acquisitions to better use shared structural information among contrasts. In the CS framework, a typical scenario involves multiple acquisitions with nearly identical acceleration rates [43]–[45]. Undersampled data are jointly processed, and a joint-sparsity regularization term improves recovery of shared features across contrasts. Another scenario involves the fully-sampled acquisition of a reference contrast that is then used as a structural prior for other contrasts [46]. Prior-guided reconstructions use regularization terms that enforce consistency of the magnitude and direction of image gradients across distinct contrasts. These previous approaches yield enhanced quality over independent processing of each contrast. However, hand-crafted regularization terms based on transforms such as total variation or wavelet reflect often suboptimal assumptions about structural similarity among separate contrasts. The proposed rsGAN method instead employs a data-driven approach to learn to utilize information from source contrast during recovery of target contrasts.

A recent study proposed a learning-based method for joint reconstruction of multi-contrast MRI data [43]. Acquisitions for separate contrasts were accelerated at identical rates. A convolutional neural network architecture was used with a subset of network weights shared across contrasts to better capture structural similarities among contrasts. While that previous method was shown to outperform conventional CS reconstructions, it is a pure reconstruction approach that can suffer from scarce sampling of high spatial frequencies at high acceleration rates. In contrast, rsGAN employs detailed structural information in a source contrast to enhance the recovery of high-frequency samples in target contrasts. Since the source acquisitions are fully-sampled or lightly undersampled, rsGAN shows improved reliability against losses in resolution. Furthermore, GANs have been shown to better learn the distribution of high-spatial-frequency information compared to conventional network architectures.

The synthesis framework is an alternative for recovery of images of a target contrast, where data are only available in a different source contrast. A powerful approach is to construct dictionaries from multi-resolution image patches, and to learn a mapping between the source and target dictionaries [15], [16], [19], [20], [24]. Segregation of the dictionary extraction and mapping stages might yield suboptimal performance. Network-based approaches offer a remedy to this problem by unifying the two stages [17], [26]–[28]. We recently proposed GAN-based synthesis for multi-contrast MRI that yielded enhanced performance compared to conventional methods [31]. Yet, due to lack of evidence on the target contrast, a pure synthesis approach can suffer from from artificial sensitivity or insensitivity to image features. The rsGAN method, on the other hand, always collects a moderate to small amount of evidence. This helps avoid artefactual feature leakage from the source to the target contrast or loss of target-contrast features that are not apparent in the source contrast.

Several technical developments are viable for improving the current implementation of the proposed method. First, the model can be generalized to simultaneously process multiple neighboring cross-sections in addition to multiple contrasts. Correlated tissue structure across cross sections might enhanced recovery despite the increase in model complexity. Second, when multiple source contrasts are present, a weight sharing method can be used to enforce a shared latent representation among contrasts for improved performance. Lastly, a cycleGAN-based model [40] might be implemented to allow for learning on unpaired multi-contrast MRI datasets that are relatively more available than paired datasets.

## 5 – Conclusion

We proposed a synergistic reconstruction-synthesis method for accelerated multi-contrast MRI based on conditional generative adversarial networks. End-to-end trained GANs are used to recover high-quality images of target and source contrasts given undersampled acquisitions. Unlike pure reconstruction, rsGAN uses high-spatial-frequency prior information in the source contrast to enhance recovery of the target contrast. Unlike pure synthesis, rsGAN bases recovered images on evidence collected through heavily undersampled acquisitions of the target contrast. The proposed method outperforms state-of-the-art reconstruction and synthesis methods, with enhanced recovery of high-frequency tissue structure, and improved reliability against feature leakage or loss. The rsGAN method holds great promise for highly accelerated multi-contrast MRI in clinical practice.

**Figures**

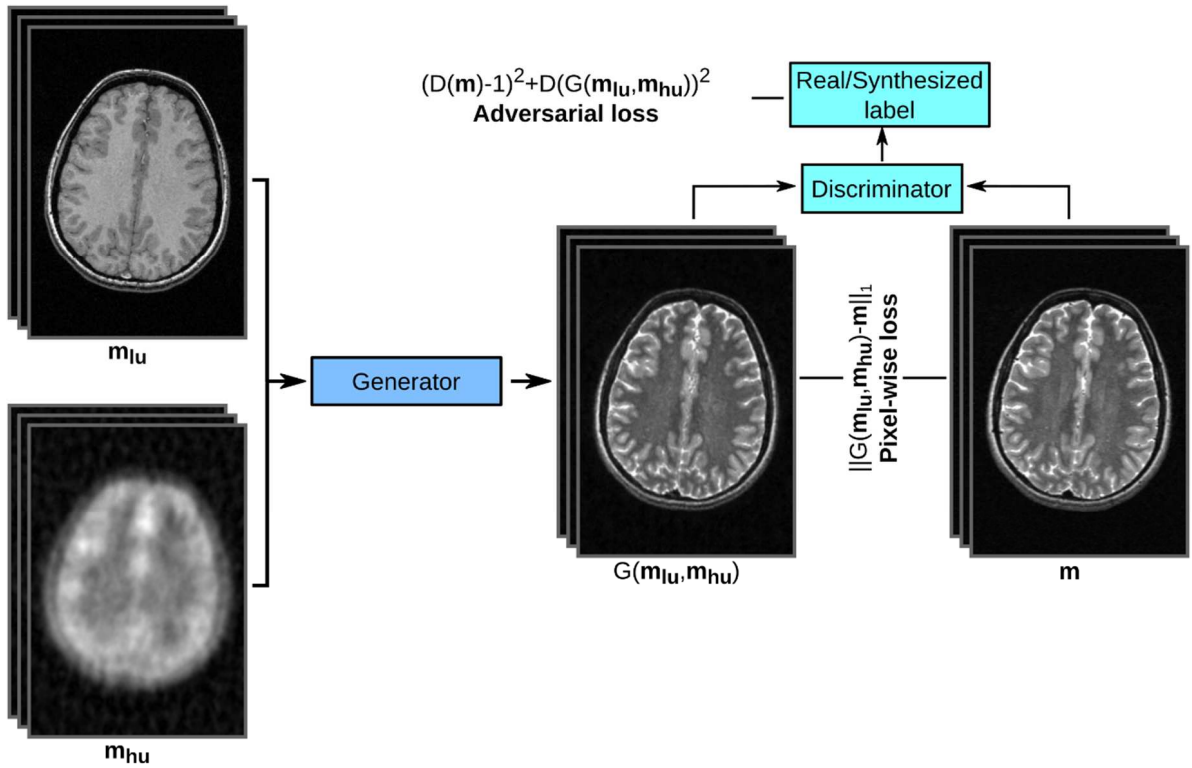

**Figure 1.** The rsGAN method is based on a conditional adversarial network with a generator $G$ and a discriminator D. Given a set of fully-sampled or lightly undersampled acquisitions of one or more source contrasts ($m_{lu}$), and highly undersampled acquisitions of one or more target contrasts ($m_{hu}$), G learns to recover realistic high-quality target-contrast images via synergistic reconstruction-synthesis. This recovery aims to minimize a pixel-wise loss function and an adversarial loss function. Meanwhile, D learns to discriminate between synthetic (G($m_{lu}$, $m_{hu}$)) and real (**m**) pairs of multi-contrast images by maximizing the adversarial loss function.

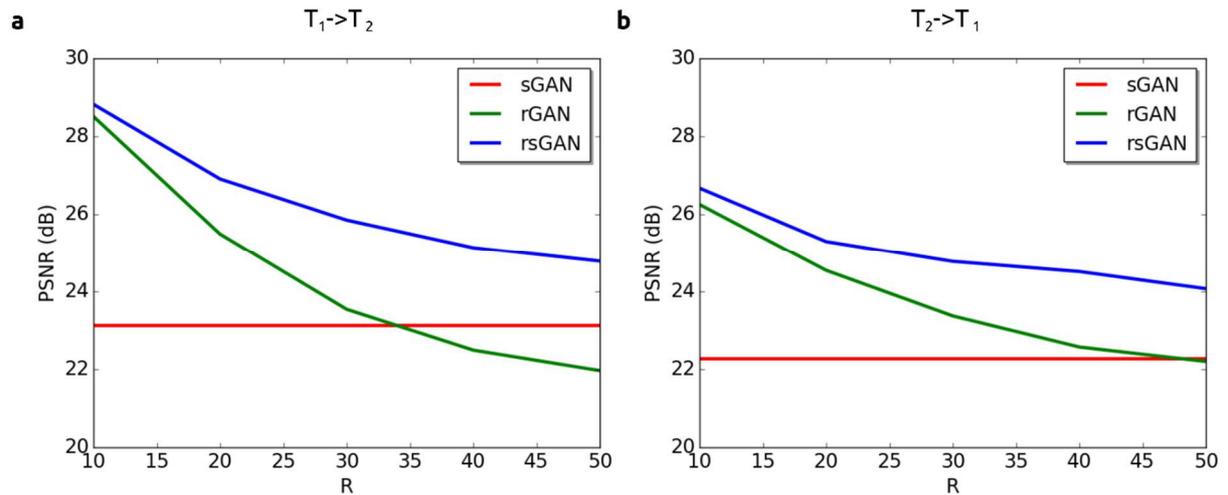

**Figure 2.** The proposed rsGAN method was demonstrated for synergistic reconstruction-synthesis of $T_1$- and $T_2$-weighted images from the MIDAS dataset. The acquisition for the source contrast was fully sampled, and the acquisition for the target contrast was undersampled by R=10x, 20x, 30x, 40x, 50x. PNSR was measured between recovered and fully-sampled reference target-contrast images. **(a)** Average PSNR across test subjects for rsGAN, rGAN and sGAN when $T_1$ is the source contrast and $T_2$ is the target contrast. **(b)** Average PSNR when $T_2$ is the source contrast and $T_1$ is the target contrast. The performance of sGAN remains constant across R since it does not use any evidence from the target-contrast acquisitions. As expected, the performance of both rGAN and rsGAN gradually decrease for higher values of R where the evidence from the target contrast becomes scarce. However, rsGAN performs well even at very high acceleration factors, and it outperforms both rGAN and sGAN across all R.

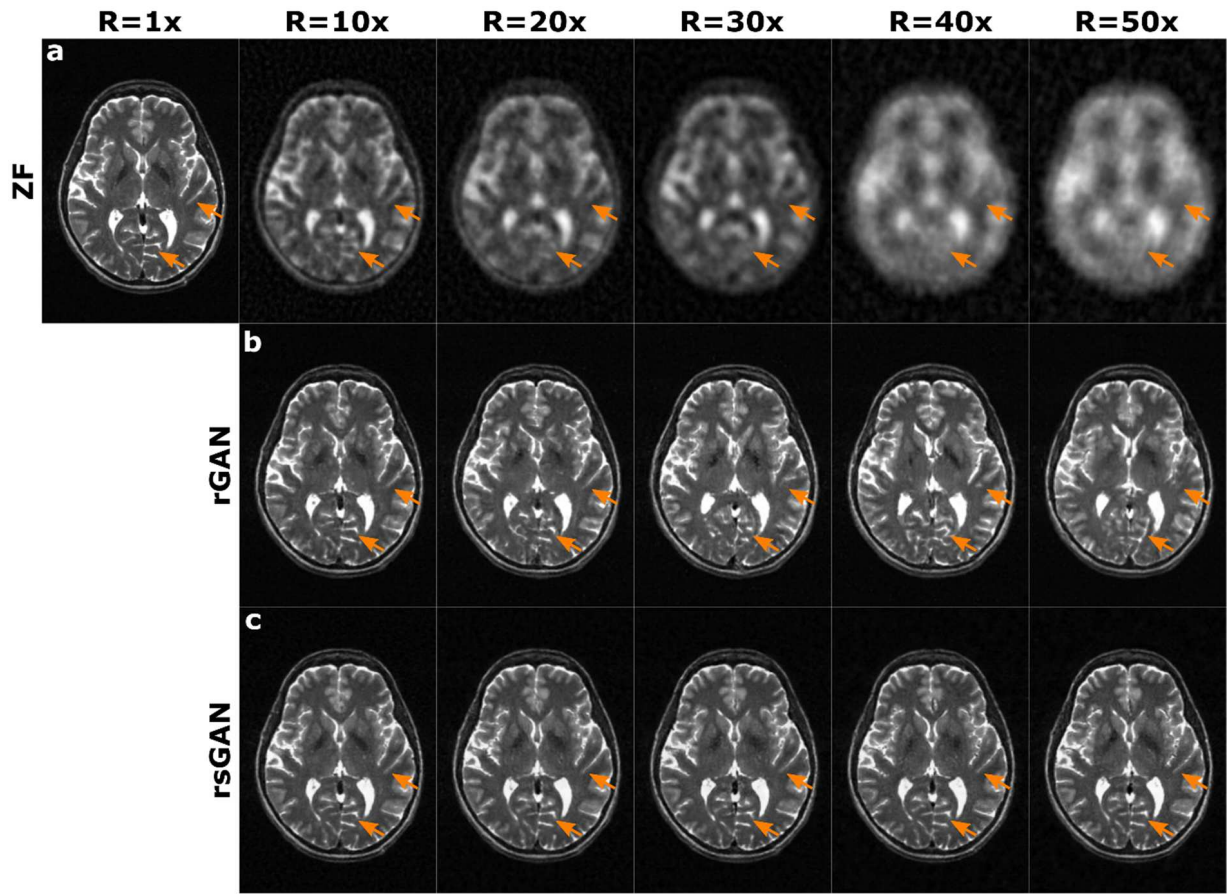

**Figure 3.** $T_2$-weighted images in the MIDAS dataset were recovered from heavily undersampled acquisitions (R=10x, 20x, 30x, 40x, 50x). The acquisition for the source contrast ($T_1$-weighted) was fully sampled. Target-contrast images recovered by **(a)** ZF (zero-filled Fourier reconstruction), **(b)** rGAN, **(c)** rsGAN. As the value of R increases the performance of rGAN degrades significantly. Meanwhile, rsGAN maintains high-quality recovered images due to use of additional information from the source contrast. Regions with enhanced recovery in rsGAN are marked with arrows.

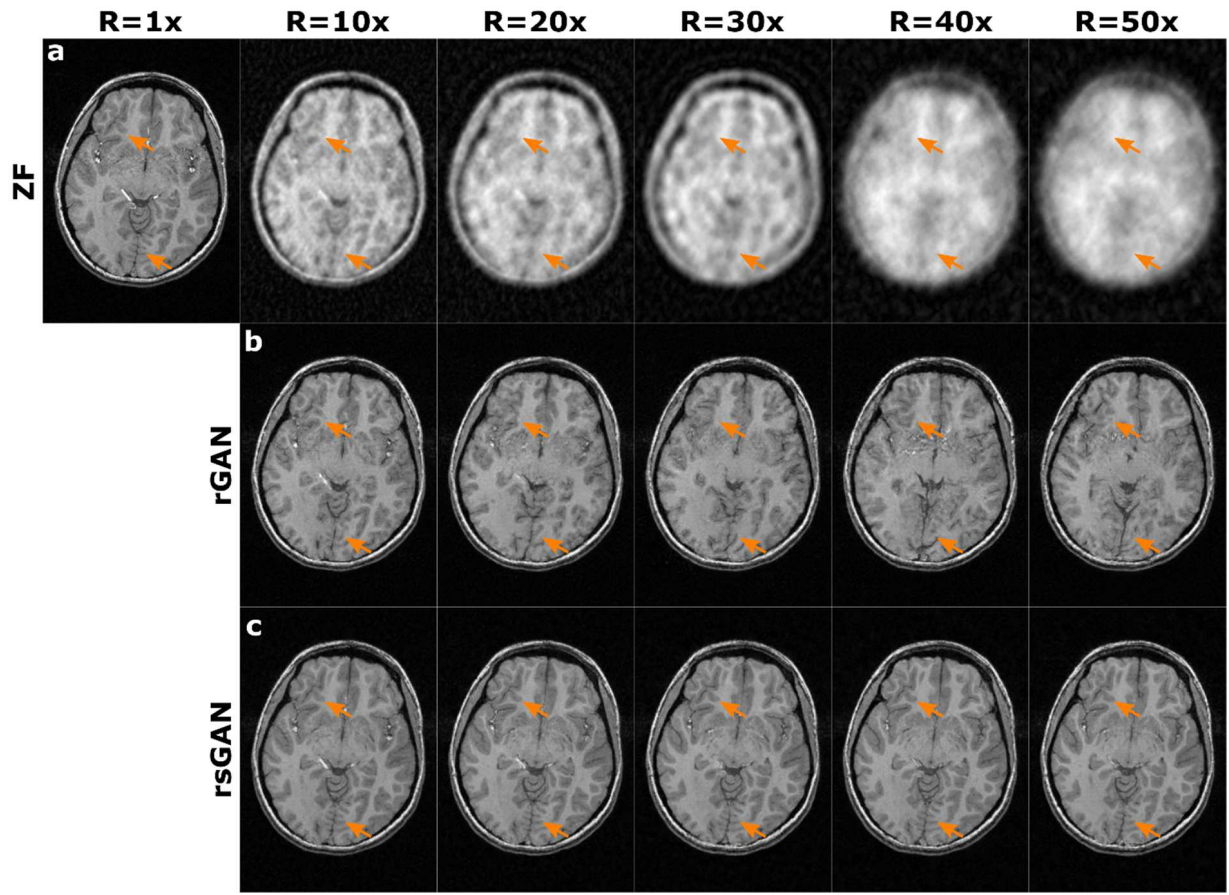

**Figure 4.** $T_1$-weighted images in the MIDAS dataset were recovered from heavily undersampled acquisitions (R=10x, 20x, 30x, 40x, 50x). The acquisition for the source contrast ($T_2$-weighted) was fully sampled. Target-contrast images recovered by **(a)** ZF, **(b)** rGAN, **(c)** rsGAN. As the value of R increases the performance of rGAN degrades significantly. Meanwhile, rsGAN maintains high-quality recovered images due to use of additional information from the source contrast. Regions with enhanced recovery in rsGAN are marked with arrows.

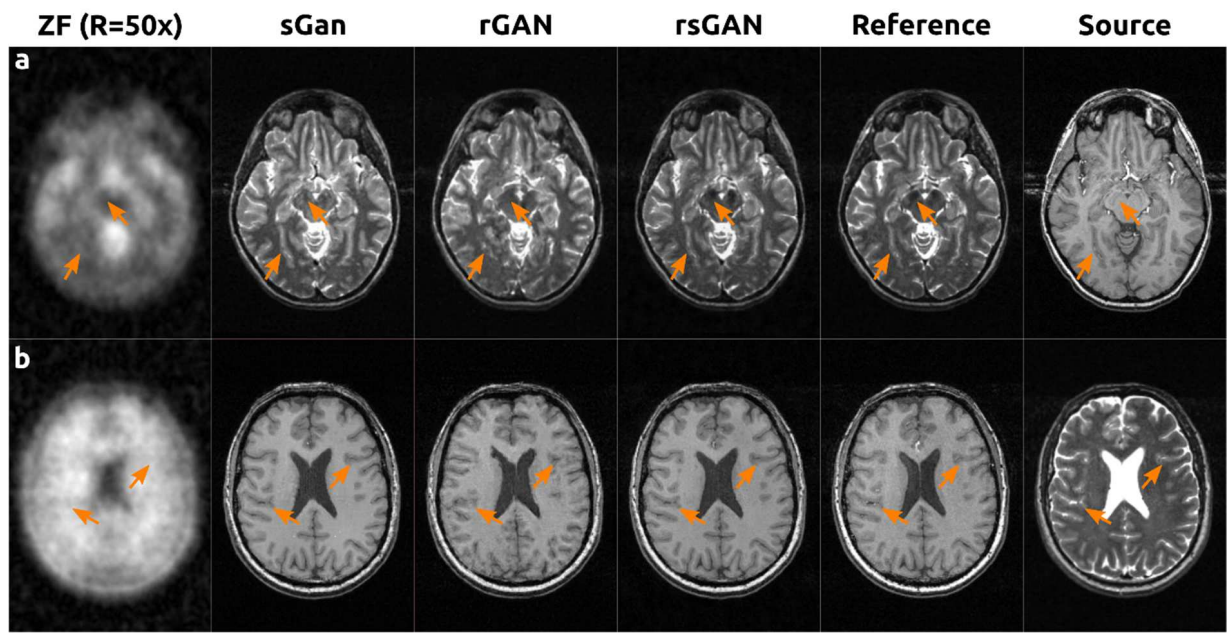

**Figure 5.** Multi-contrast images in the MIDAS dataset were recovered, where the source contrast was fully sampled and the target contrast was undersampled at R=50x. Images were recovered using ZF, sGAN, rGAN and rsGAN. **(a)** Recovered $T_2$-weighted images are shown along with the fully-sampled reference image and the source-contrast image. **(b)** Recovered $T_1$-weighted images are shown along with the fully-sampled reference image and the source-contrast image. rsGAN yields visually accurate recovery of the target-contrast image compared to sGAN and rGAN. Sample regions that are better recovered by rsGAN are marked with arrows.

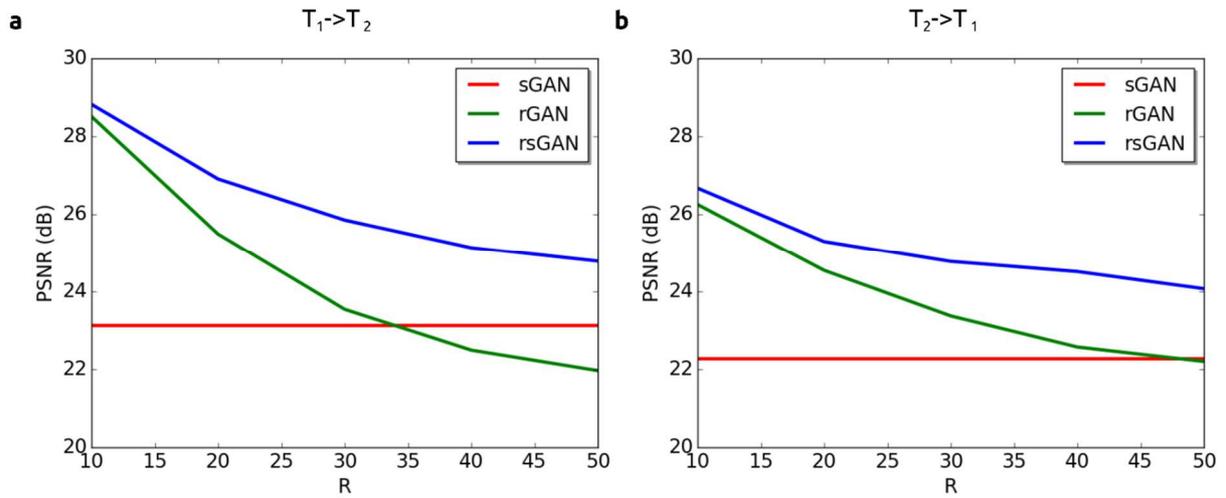

**Figure 6.** The proposed rsGAN method was demonstrated for synergistic reconstruction-synthesis of $T_1$- and $T_2$-weighted images from the BRATS dataset. The acquisition for the source contrast was fully sampled, and the acquisition for the target contrast was undersampled by R=10x, 20x, 30x, 40x, 50x. PNSR was measured between recovered and fully-sampled reference target-contrast images. **(a)** Average PSNR for rsGAN, rGAN and sGAN when $T_1$ is the source contrast and $T_2$ is the target contrast. **(b)** Average PSNR when $T_2$ is the source contrast and $T_1$ is the target contrast. As expected, the performance of sGAN remains constant across R, whereas the performances of rGAN and rsGAN gradually decrease for higher values of R. Meanwhile, rsGAN outperforms rGAN at all R, and the performance difference between the two methods increases for higher R. The rsGAN method also outperforms sGAN in all cases.

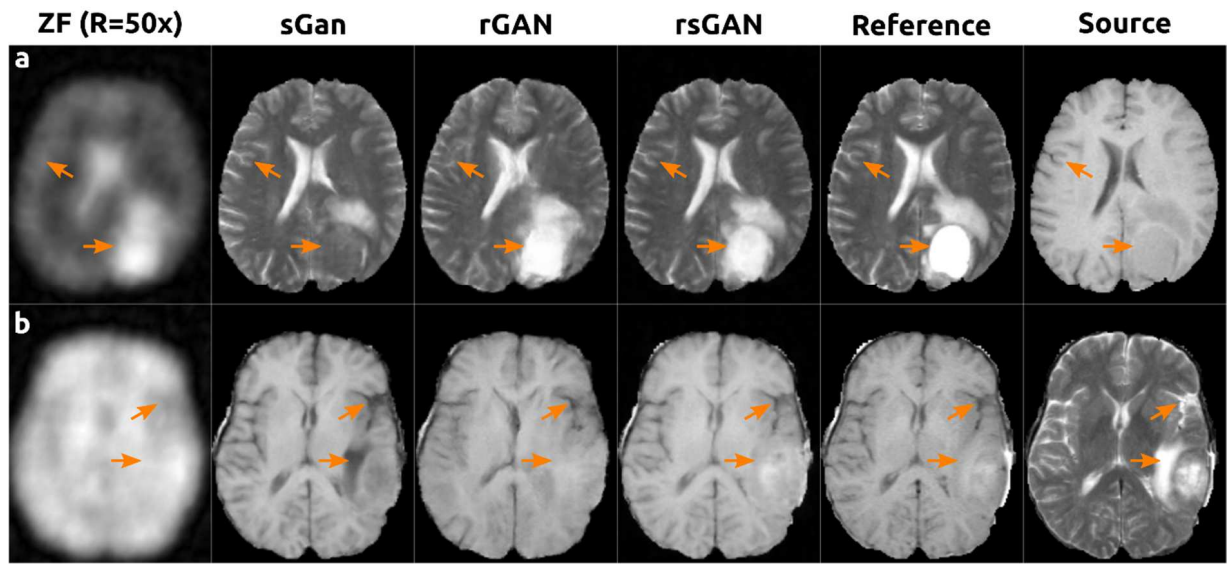

**Figure 7.** Multi-contrast images in the BRATS dataset were recovered, where the source contrast was fully sampled and the target contrast was undersampled at R=50x. Images were recovered using ZF, sGAN, rGAN and rsGAN. (**a**) Recovered $T_2$-weighted images along with the fully-sampled reference image and the source-contrast image. (**b**) Recovered $T_1$-weighted images along with the fully-sampled reference image and the source-contrast image. rsGAN yields visually superior images compared to sGAN and rGAN. Note that sGAN suffers from either loss of features in the target contrast (in **a**) or synthesis of artefactual features (in **b**). Meanwhile, rGAN suffers from excessive loss of high spatial frequency information. Sample regions that are more accurately recovered by rsGAN are marked with arrows.

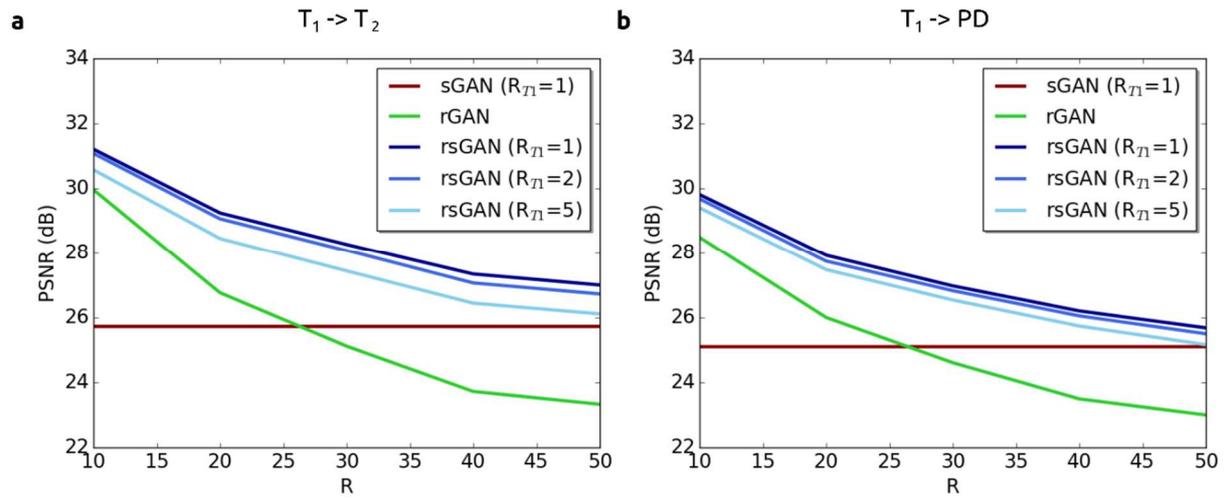

**Figure 8.** The proposed rsGAN method was demonstrated for synergistic reconstruction-synthesis of $T_1$-, $T_2$- and PD-weighted images from the BRATS dataset. The acquisition for the source contrast ($T_1$-weighted) was lightly undersampled by $R_{T1}$=1x, 2x, 5x, and the acquisitions for the target contrasts ($T_2$- and PD-weighted) were heavily undersampled by R=10x, 20x, 30x, 40x, 50x. **(a)** Average PSNR on $T_2$-weighted images across test subjects for sGAN, rGAN, and rsGAN. **(b)** Average PSNR on $T_1$-weighted images for sGAN, rGAN, and rsGAN. As expected, rsGAN outperforms both sGAN and rGAN at all R. At the same time, performance of rsGAN is highly similar for distinct values of $R_{T1}$.

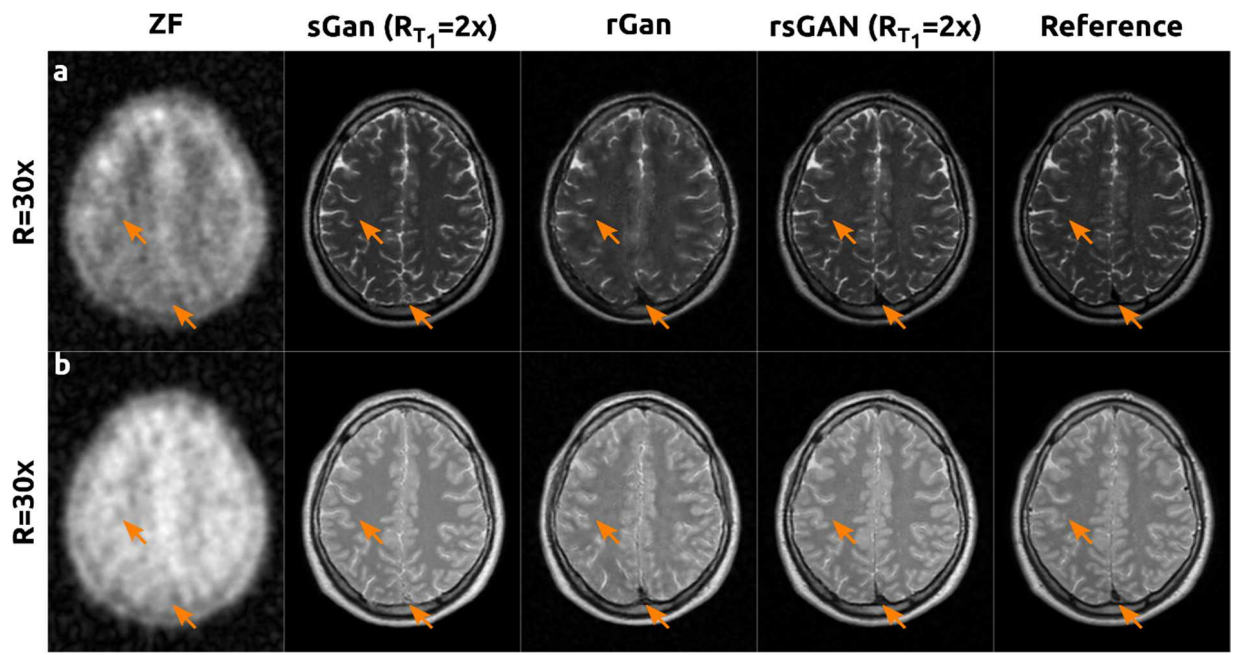

**Figure 9.** Multi-contrast images in the IXI dataset were recovered, where the source contrast ($T_1$-weighted) was lightly undersampled at $R_{T_1}=2$x, and the target contrasts ($T_2$- and PD-weighted) were heavily undersampled at R=30x. Images were recovered using ZF, sGAN, rGAN and rsGAN. **(a)** Recovered $T_2$-weighted images. **(b)** Recovered PD-weighted images. Samples regions where rsGAN yields sharper images and improved suppression of aliasing artifacts are marked with arrows.

**TABLES**

**TABLE I** – Quality of recovered images in the MIDAS dataset. PSNR and SSIM values across the test subjects are listed for sGAN, rGAN, and rsGAN. $T_1$-weighted acquisitions were taken as the source contrast, and $T_2$-weighted acquisitions were taken as the target contrast. The highest PSNR and SSIM values in each row are marked in bold font (p<0.01).

|         | sGAN          |               | rGAN          |               | rsGAN         |               |
|---------|---------------|---------------|---------------|---------------|---------------|---------------|
|         | PSNR          | SSIM          | PSNR          | SSIM          | PSNR          | SSIM          |
| R = 10x | 23.06 ± 0.53  | 0.862 ± 0.016 | **28.62 ± 0.27** | 0.928 ± 0.006 | **28.80 ± 0.39** | **0.935 ± 0.008** |
| R = 20x | 23.06 ± 0.53  | 0.862 ± 0.016 | 25.67 ± 0.29  | 0.885 ± 0.007 | **26.92 ± 0.30** | **0.912 ± 0.009** |
| R = 30x | 23.06 ± 0.53  | 0.862 ± 0.016 | 23.75 ± 0.36  | 0.850 ± 0.009 | **25.94 ± 0.32** | **0.902 ± 0.009** |
| R = 40x | 23.06 ± 0.53  | 0.862 ± 0.016 | 22.70 ± 0.36  | 0.821 ± 0.010 | **25.15 ± 0.31** | **0.886 ± 0.010** |
| R = 50x | 23.06 ± 0.53  | **0.862 ± 0.016** | 22.16 ± 0.38 | 0.807 ± 0.010 | **24.76 ± 0.47** | **0.875 ± 0.020** |

**TABLE II** – Quality of recovered images in the MIDAS dataset. PSNR and SSIM values across the test subjects are listed for sGAN, rGAN, and rsGAN. $T_2$-weighted acquisitions were taken as the source contrast, and $T_1$-weighted acquisitions were taken as the target contrast. The highest PSNR and SSIM values in each row are marked in bold font ($p<0.01$).

|  | sGAN | | rGAN | | rsGAN | |
|---|---|---|---|---|---|---|
|  | PSNR | SSIM | PSNR | SSIM | PSNR | SSIM |
| R = 10x | 22.03 ± 0.93 | 0.795 ± 0.027 | 26.26 ± 0.49 | 0.866 ± 0.016 | **26.62 ± 0.47** | **0.881 ± 0.016** |
| R = 20x | 22.03 ± 0.93 | 0.795 ± 0.027 | 24.62 ± 0.45 | **0.825 ± 0.017** | **25.15 ± 0.57** | **0.838 ± 0.022** |
| R = 30x | 22.03 ± 0.93 | 0.795 ± 0.027 | 23.48 ± 0.45 | 0.791 ± 0.019 | **24.72 ± 0.48** | **0.848 ± 0.019** |
| R = 40x | 22.03 ± 0.93 | 0.795 ± 0.027 | 22.71 ± 0.42 | 0.763 ± 0.018 | **24.45 ± 0.48** | **0.841 ± 0.018** |
| R = 50x | 22.03 ± 0.93 | 0.795 ± 0.027 | 22.36 ± 0.44 | 0.751 ± 0.019 | **23.99 ± 0.53** | **0.829 ± 0.017** |

**TABLE III** – Quality of recovered images in the BRATS dataset. PSNR and SSIM values across the test subjects are listed for sGAN, rGAN, and rsGAN. $T_1$-weighted acquisitions were taken as the source contrast, and $T_2$-weighted acquisitions were taken as the target contrast. The highest PSNR and SSIM values in each row are marked in bold font ($p<0.01$).

|  | sGAN | | rGAN | | rsGAN | |
|---|---|---|---|---|---|---|
|  | PSNR | SSIM | PSNR | SSIM | PSNR | SSIM |
| R = 10x | 20.91 ± 1.09 | 0.832 ± 0.036 | **29.84 ± 1.15** | **0.918 ± 0.012** | **29.98 ± 1.26** | **0.919 ± 0.017** |
| R = 20x | 20.91 ± 1.09 | 0.832 ± 0.036 | **27.22 ± 0.96** | **0.867 ± 0.016** | **27.72 ± 1.03** | **0.879 ± 0.025** |
| R = 30x | 20.91 ± 1.09 | 0.832 ± 0.036 | 25.57 ± 0.90 | 0.835 ± 0.017 | **26.50 ± 1.06** | **0.866 ± 0.029** |
| R = 40x | 20.91 ± 1.09 | **0.832 ± 0.036** | 23.39 ± 0.99 | 0.761 ± 0.025 | **25.61 ± 1.03** | **0.839 ± 0.035** |
| R = 50x | 20.91 ± 1.09 | **0.832 ± 0.036** | 23.65 ± 0.73 | 0.751 ± 0.023 | **25.45 ± 1.03** | **0.825 ± 0.038** |

**TABLE IV** – Quality of recovered images in the BRATS dataset. PSNR and SSIM values across the test subjects are listed for sGAN, rGAN, and rsGAN. $T_2$-weighted acquisitions were taken as the source contrast, and $T_1$-weighted acquisitions were taken as the target contrast. The highest PSNR and SSIM values in each row are marked in bold font (p<0.01).

|  | sGAN | | rGAN | | rsGAN | |
|---|---|---|---|---|---|---|
|  | PSNR | SSIM | PSNR | SSIM | PSNR | SSIM |
| R = 10x | 20.90 ± 1.84 | 0.869 ± 0.031 | 28.36 ± 1.44 | **0.922 ± 0.008** | **29.11 ± 1.48** | **0.930 ± 0.013** |
| R = 20x | 20.90 ± 1.84 | 0.869 ± 0.031 | **26.53 ± 0.99** | 0.884 ± 0.007 | **27.51 ± 1.42** | **0.905 ± 0.020** |
| R = 30x | 20.90 ± 1.84 | 0.869 ± 0.031 | 24.27 ± 1.14 | 0.846 ± 0.011 | **26.78 ± 1.20** | **0.899 ± 0.024** |
| R = 40x | 20.90 ± 1.84 | **0.869 ± 0.031** | 23.13 ± 1.08 | 0.792 ± 0.017 | **25.41 ± 1.39** | **0.869 ± 0.034** |
| R = 50x | 20.90 ± 1.84 | **0.869 ± 0.031** | 22.45 ± 1.23 | 0.775 ± 0.019 | **24.82 ± 1.39** | **0.855 ± 0.038** |

**TABLE V** – Quality of recovered images in the IXI dataset. $T_1$-weighted acquisitions accelerated to various degrees ($R_{T1}$) were taken as the source contrast, and $T_2$- and PD-weighted acquisitions were taken as the target contrasts. PSNR and SSIM values for $T_2$-weighted images across the test subjects are listed for sGAN, rGAN, and rsGAN. The highest PSNR and SSIM values in each row are marked in bold font (p<0.01).

|  | sGAN ($R_{T1}$=1) | | rGAN | | rsGAN ($R_{T1}$=1) | | rsGAN ($R_{T1}$=2) | | rsGAN ($R_{T1}$=5) | |
|---|---|---|---|---|---|---|---|---|---|---|
|  | PSNR | SSIM | PSNR | SSIM | PSNR | SSIM | PSNR | SSIM | PSNR | SSIM |
| R = 10x | 25.73 ± 1.04 | 0.905 ± 0.017 | 29.96 ± 0.71 | 0.881 ± 0.009 | **31.21 ± 0.85** | **0.915 ± 0.007** | 31.09 ± 0.87 | 0.912 ± 0.008 | 30.58 ± 0.87 | 0.901 ± 0.008 |
| R = 20x | 25.73 ± 1.04 | **0.905 ± 0.017** | 26.76 ± 0.80 | 0.818 ± 0.015 | **29.25 ± 0.88** | **0.896 ± 0.011** | 29.07 ± 0.86 | 0.892 ± 0.010 | 28.45 ± 0.85 | 0.878 ± 0.012 |
| R = 30x | 25.73 ± 1.04 | **0.905 ± 0.017** | 25.11 ± 0.87 | 0.798 ± 0.022 | **28.27 ± 0.85** | **0.901 ± 0.011** | 28.08 ± 0.82 | 0.896 ± 0.010 | 27.44 ± 0.81 | 0.881 ± 0.012 |
| R = 40x | 25.73 ± 1.04 | **0.905 ± 0.017** | 23.72 ± 0.81 | 0.741 ± 0.026 | **27.35 ± 0.91** | 0.882 ± 0.016 | 27.07 ± 0.88 | 0.874 ± 0.015 | 26.44 ± 0.85 | 0.858 ± 0.017 |
| R = 50x | 25.73 ± 1.04 | **0.905 ± 0.017** | 23.32 ± 0.86 | 0.716 ± 0.028 | **27.01 ± 0.97** | 0.867 ± 0.018 | 26.73 ± 0.92 | 0.860 ± 0.017 | 26.11 ± 0.83 | 0.843 ± 0.017 |

**TABLE VI** – Quality of recovered images in the IXI dataset. $T_1$-weighted acquisitions accelerated to various degrees ($R_{T1}$) were taken as the source contrast, and $T_2$- and PD-weighted acquisitions were taken as the target contrasts. PSNR and SSIM values for PD-weighted images across the test subjects are listed for sGAN, rGAN and rsGAN. The highest PSNR and SSIM values in each row are marked in bold font (p<0.01).

|  | sGAN ($R_{T1}$=1) | | rGAN | | rsGAN ($R_{T1}$=1) | | rsGAN ($R_{T1}$=2) | | rsGAN ($R_{T1}$=5) | |
| --- | --- | --- | --- | --- | --- | --- | --- | --- | --- | --- |
|  | PSNR | SSIM | PSNR | SSIM | PSNR | SSIM | PSNR | SSIM | PSNR | SSIM |
| R = 10x | 25.10 ± 1.17 | **0.918 ± 0.015** | 28.48 ± 1.37 | 0.877 ± 0.014 | **29.82 ± 1.11** | **0.917 ± 0.008** | 29.69 ± 1.16 | 0.915 ± 0.008 | 29.41 ± 1.15 | 0.907 ± 0.009 |
| R = 20x | 25.10 ± 1.17 | **0.918 ± 0.015** | 26.00 ± 1.35 | 0.805 ± 0.023 | **27.94 ± 1.11** | 0.899 ± 0.010 | 27.76 ± 1.18 | 0.894 ± 0.010 | 27.50 ± 1.14 | 0.884 ± 0.012 |
| R = 30x | 25.10 ± 1.17 | **0.918 ± 0.015** | 24.61 ± 1.27 | 0.789 ± 0.028 | **27.00 ± 1.11** | 0.900 ± 0.011 | 26.85 ± 1.09 | 0.895 ± 0.011 | 26.56 ± 01.05 | 0.885 ± 0.013 |
| R = 40x | 25.10 ± 1.17 | **0.918 ± 0.015** | 23.50 ± 1.18 | 0.737 ± 0.036 | **26.21 ± 1.08** | 0.883 ± 0.015 | 26.05 ± 1.05 | 0.876 ± 0.014 | 25.75 ± 1.02 | 0.865 ± 0.017 |
| R = 50x | **25.10 ± 1.17** | **0.918 ± 0.015** | 23.00 ± 1.12 | 0.707 ± 0.033 | **25.70 ± 0.99** | 0.867 ± 0.017 | **25.52 ± 1.00** | 0.859 ± 0.018 | **25.18 ± 0.97** | 0.846 ± 0.019 |